\documentclass[conference]{IEEEtran}
\IEEEoverridecommandlockouts
\usepackage{cite}
\usepackage{amsmath,amssymb,amsfonts}
\usepackage{algorithmic}
\usepackage{graphicx}
\usepackage{textcomp}
\usepackage{xcolor}
\def\BibTeX{{\rm B\kern-.05em{\sc i\kern-.025em b}\kern-.08em
    T\kern-.1667em\lower.7ex\hbox{E}\kern-.125emX}}
\begin{document}

\title{Fortify Machine Learning Production Systems:\\ Detect and Classify Adversarial Attacks}

\author{\IEEEauthorblockN{Matthew Ciolino}
\IEEEauthorblockA{\textit{PeopleTec Inc.} \\
Huntsville, AL, USA \\
matt.ciolino@peopletec.com}
\and
\IEEEauthorblockN{Josh Kalin}
\IEEEauthorblockA{\textit{Department of Computer Science} \\
\textit{Auburn University}\\
Auburn, AL, USA \\
josh.kalin@peopletec.com}
\and
\IEEEauthorblockN{David Noever}
\IEEEauthorblockA{\textit{PeopleTec Inc.} \\
Huntsville, AL, USA \\
david.noever@peopletec.com}
}

\maketitle

\begin{abstract}
Production machine learning systems are consistently under attack by adversarial actors. Various deep learning models must be capable of accurately detecting fake or adversarial input  while maintaining speed. In this work, we propose one piece of the production protection system: detecting an incoming adversarial attack and its characteristics. Detecting types of adversarial attacks has two primary effects: the underlying model can be trained in a structured manner to be robust from those attacks and the attacks can be potentially filtered out in real time before causing any downstream damage. The adversarial image classification space is explored for models commonly used in transfer learning.
\end{abstract}


\section{Introduction}
In our previous work \cite{kalin2020black} we attempted to find machine learning model characteristics through probing black box models. In that work, our goal was to classify the classifiers used. We looked at three stages of strategic discovery: probe, collect, and detect. On the image side, we would probe by sending in a picture, collect the model output (1000d vector for ImageNet), and using that output and the model name, trained a new classifier. We could now tell if a model output came from ResNet50, DenseNet121, etc. at 0.99 average precision (AP). 

We took this one step forward and fine-tuned MobileNetV2 on different 50-class datasets. We then ran inference from those fine-tuned models on a mix of super resolution datasets and collected the model output alongside the fine-tuned model's name. We trained a classifier and we were able to predict which dataset MobileNetV2 was trained on based on a single model output at 0.97 AP. 

We were successfully able to classify both which model the prediction came from alongside the dataset that the model was fine-tuned on. Extending this work, we now introduce the adversarial version in this paper. We attempt to detect if a model was attacked and the attacker's characteristics. This defensive step completes the analysis of the vulnerable surface of a machine learning model. We can detect which dataset was used to train a model, which model was used, if an attack occurred, and finally which attack was used, all from a single model output.

\subsection{Background}
The adversarial attack surface \cite{huang2011adversarial} \cite{machado2020adversarial} \cite{chakraborty2018adversarial} \cite{xu2020adversarial} represents all of the ways an attacker can manipulate or avoid detection from machine learning systems. As machine learning models are incorporated to more production systems, developers  need to find ways to harden and strengthen those systems from adversarial actors. Adversarial attacks will exploit different qualities about a machine learning system - typically, focusing on vulnerabilities discovered from either the dataset or the model architecture. 

 Wang, et. al \cite{wang2019security} outlines seven different adversarial attack concepts encountered during the machine learning training process. In our paper, we focus on Learning Algorithm (attacked by) Gradient Descent Attack (among other). Wang, et. al asserts, "it is observed that ML methods are highly vulnerable to adversarial attacks during both the training and prediction phases [and] DNNs [deep neural networks] are vulnerable to subtle input perturbations that result in substantial changes in outputs." 

A novel approach to detect adversarial attack comes from using Trapdoors in a feature embedding of convolutional networks as discussed by Shan et. al \cite{shan2019using}. Instead inspecting the output of a model, they attempt to discover an adversarial attack during inference. They use the optimization of adversarial algorithms against the attacker by placing trapdoors that attempt to shortcut the attack from one class to another (y$_t$ to y$_x$). This leads the attacker to a false local minimum while reducing the effect on performance on the rest of the network.

Another approach which couples a model to the original network is proposed by Metzen et. al \cite{metzen2017detecting}. Metzen grabs feature embeddings at various depths in the network and uses a subnet to classify if the model is being attacked. He suggests that this could be used as a mitigating factor during inference to block adversarial actors.

\subsection{Contributions}
While the above methods couple to a single machine learning model, our novel approach attempts to create a universal detector for adversarial attacks and their characteristics. In our previous work, we demonstrated the ability of an attacker to detect the underlying dataset and model architecture from strategic probing of the machine learning system. This work  demonstrably shows the defense against adversarial attacks on image classification models: We show (i) the effectiveness of common white box adversarial attacks, (ii) that we can detect whether a model output is the results of an adversarial attack (i.e. the perturbation of the attack is too strong), (iii) that we can detect from a model's output, which model the adversarial method attacked, and (iv) from a model's output, the attack method used.

\subsection{Reproducability}

To validate the experiment we share a 
\textbf{Reproducible Google Collaboratory \cite{reproducible_colab}} which can run a subset of the dataset used.

\section{Experiment}

We apply six (6) different adversarial attacks to five (5) different pre-trained models. We use the ImageNetV2 \cite{recht2019imagenet} test dataset and attempt to answer four (4) questions using sklearn's \cite{scikit-learn} standard Random Forest Algorithm. The experimental diagram is proposed in [Figure \ref{exp_flow}]:

\begin{figure}[t]
    \centering
    \includegraphics[scale=.4]{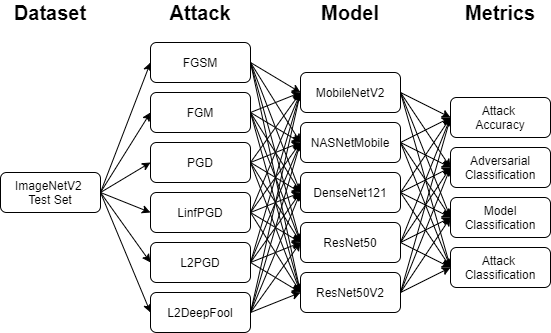}
    \caption{Adversarial Image Experiment}
    \label{exp_flow}
\end{figure}

\subsection{Adversarial Attacks}

We begin by describing the six different white box attacks from the adversarial attack package, Foolbox \cite{rauber2017foolboxnative}. In this experiment: \textbf{FGM, FGSM, PGD, LinfPGD, L2PGD, L2DeepFool} \medskip

\begin{quote}
\textbf{Gradient Sign Attack (FGSM)}

Computes the gradient $g(x_0) = \nabla_{x} L(x_0,\ell_0)$ once and then seeks the minimum step size $\varepsilon$ such that $x_0 + \varepsilon sign(g(x_0))$ is adversarial. (Goodfellow et al., 2014). (Rauber et al. 2018 \cite{rauber2017foolbox})\\
\end{quote} 

\begin{quote}
\textbf{Gradient Sign Attack (FGM)}

Extends the FGSM to other norms [from infinity norm] and is therefore called the Fast Gradient Method. (adversarial-robustness-toolbox \cite{art2018})\\
\end{quote}

\begin{quote}
\textbf{DeepFool L2 Attack}

In each iteration DeepFool (Moosavi-Dezfooli et al., 2015)
computes for each class $\ell \neq \ell_0$ the minimum distance $d(\ell,\ell_0)$ that it takes to reach the class boundary by approximating the model classifier with a linear classifier. It then
makes a corresponding step in the direction of the class
with the smallest distance. (Rauber et al. 2018 \cite{rauber2017foolbox})\\
\end{quote}

\begin{quote}
\textbf{Projected Gradient Descent (PGD)}

Projected Gradient Descent (PGD) \cite{madry2017towards} is also an iterative extension of FGSM and very similar to Basic Iterative Method \cite{kurakin2016adversarial} (BIM). The main difference with BIM resides in the fact that PGD projects the attack result back on the $\varepsilon$-norm ball around the original input at each iteration of the attack. (adversarial-robustness-toolbox \cite{art2018})\\
\end{quote}

\begin{quote}
\textbf{Linf Projected Gradient Descent (LinfPGD)} \\
Linf Projected Gradient Descent is a PGD attack with order = Linf. (Foolbox \cite{rauber2017foolboxnative})\\
\end{quote}

\begin{quote}
\textbf{L2 Projected Gradient Descent (L2PGD)} \\
L2 Projected Gradient Descent is a PGD attack with order = L2. (Foolbox \cite{rauber2017foolboxnative})\\
\end{quote}

The adversarial attacks listed above have the following hyper parameters: FGM (Epsilon:5), FGSM (Epsilon:.03), PGD (Epsilon:.03, Steps:10), LinfPGD (Epsilon:.1, Steps:10), L2PGD (Epsilon:5, Steps:10), L2DeepFool (Epsilon:5, Steps:10).

\subsection{Pre-trained Models}

The pre-trained models used are from keras.applications and vary in input, performance, and model size. The list [Table \ref{pretrained_models}] contains:
\textbf{MobileNetV2, NASNetMobile, DenseNet121, ResNet50, ResNet50V2}\\

\begin{table}[t]
\centering
\caption{Pre-trained Model Parameters}
    \begin{tabular}{|c|c|c|c|c|c|}
    \hline
    \textbf{Model} & \textbf{Input} & \textbf{Top-1} & \textbf{Top-5} & \textbf{Size} \\ \hline
    MobileNetV2 & 224 & 71.336 & 90.142 & 3.5M \\ \hline
    NASNetMobile & 224 & 74.366 & 91.854 & 7.7M \\ \hline
    DenseNet121 & 224 & 74.972 & 92.258 & 8.1M \\ \hline
    ResNet50 & 224 & 74.928 & 92.060 & 25.6M \\ \hline
    ResNet50V2 & 299 & 75.960 & 93.034 & 25.6M \\ \hline
    \end{tabular}
    \label{pretrained_models}
\end{table}

\subsection{Metrics} 

To analyze if we can detect a model is being attacked, at a universal level, we look at just the prediction itself. 

\begin{enumerate}
  \item \textbf{How well does the adversarial attack perform on this model and dataset?}
  \item \textbf{Once a prediction is made, can we tell if it is the result of an adversarial input or not?}
  \item \textbf{Once a prediction is made, can we tell what model the prediction came from after an adversarial attack?}
  \item \textbf{Once a prediction is made, which attack method was used?}
\end{enumerate}

\section{Evaluation}

We randomly select 250 images of the ImageNetV2 test dataset and create 250 adversarial images for each of the six attacks for each of the five models (7500 total). We run the clean images and the adversarial images through the model and collect the output. The following contains the data used for the experiments: \textbf{the sample images, the adversarial images, the sample prediction, the adversarial prediction, the truth label, a flag for if the attack was successful, the pre-trained model attacked, and the attack type.} We can then use that data to predict the metrics described above. We first start with the accuracy of the adversarial attacks [Figure \ref{attack_accuracy}] on the pre-trained models.

\subsection{\textbf{Adversarial Attack Accuracy}}

\begin{figure}[t]
    \centering
    \includegraphics[width=0.5\textwidth]{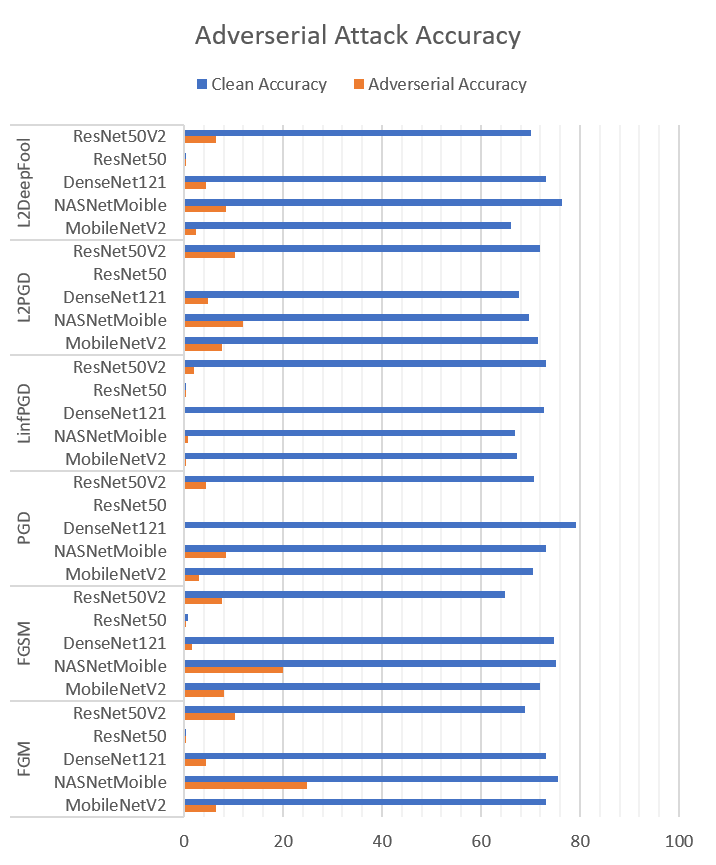}
    \caption{Original model accuracy and adversarial input accuracy for the six attacks on the five models.}
    \label{attack_accuracy}
\end{figure}

\begin{center}
    \begin{table}[t]
    \caption{Adversarial Attack Accuracy}
    \centering
        \begin{tabular}{|c|c|c|c|c|}
        \hline 
        \textbf{Model} & \textbf{Clean Avg} & \textbf{Adv. Avg} & \textbf{Clean Std} & \textbf{Adv. Std}\\ \hline 
        MobileNetV2	 & 70.07	& 4.63	& 2.69	& 3.36 \\ \hline 
        NASNetMoible & 72.80	& 12.40	& 3.99	& 6.96 \\ \hline 
        DenseNet121	 & 73.47	& 2.53	& 4.17	& 2.33 \\ \hline 
        ResNet50	 & 0.33	    & 0.27	& 0.33	& 0.22 \\ \hline 
        ResNet50V2	 & 69.93	& 6.87	& 3.23	&3.18 \\ \hline
        \multicolumn{4}{l}{$^{\mathrm{a}}$Figure \ref{attack_accuracy} Statistics}
        \end{tabular}
    \label{attack_accuracy_table}
    \end{table}
\end{center}

On average the models performed at above 70\% accuracy on the sample ImageNetV2 images [Figure \ref{attack_accuracy}, Table \ref{attack_accuracy_table}]. For the adversarial attacks the accuracy was on average below 10\%. None of the proposed attacks had any affected on the Residual Network (ResNet50) while having an effect on the ResNet50V2 network. Discussion of this anomaly is present in the discussion session [\ref{resnet50_discussion}]. ResNet50 is excluded from the next parts of the experiment. Most of the white box attacks on the five models have little variance besides for NASNetMobile. Overall, the attacks were successful and will lead to a good downstream performance in this experiment. \\

\subsection{\textbf{Adversarial Input}}

\begin{figure}[t]
    \centering
    \includegraphics[width=0.5\textwidth]{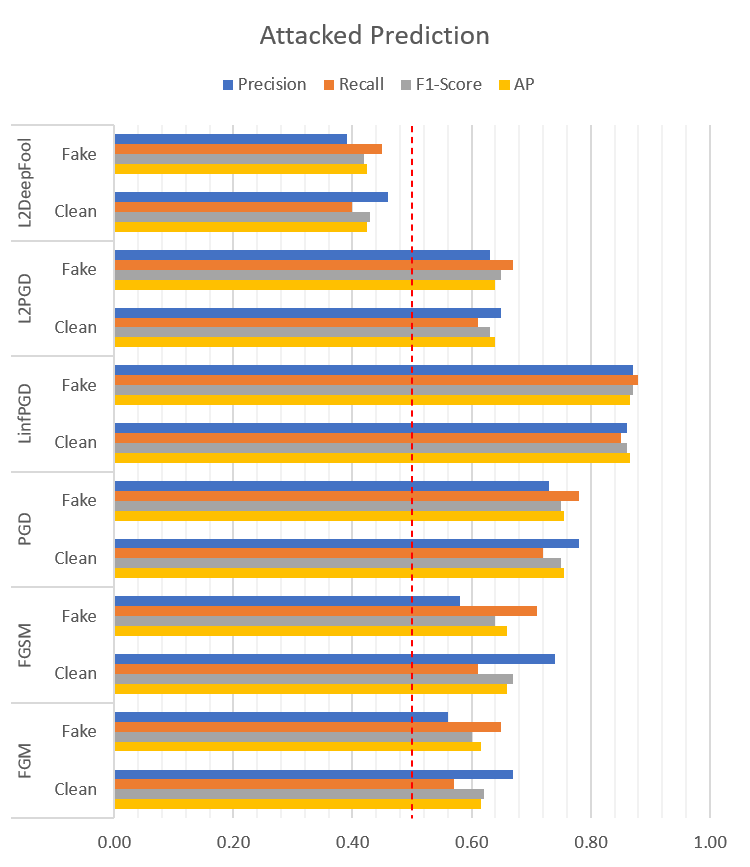}
    \caption{Binary classification of fake or clean input images. Vertical dashed red line is random guessing.}
    \label{attacked_model}
\end{figure}

\begin{center}
    \begin{table}[t]
    \caption{Attack made based on model prediction} 
    \centering
        \begin{tabular}{|c|c|c|c|}
        \hline 
        \textbf{Image Type} & \textbf{mAP} & \textbf{Average Recall} & \textbf{Average F1} \\\hline 
        Clean	& 0.693	& 0.627	& 0.660 \\\hline 
        Adversarial	& 0.627	& 0.690	& 0.655 \\\hline
        \multicolumn{4}{l}{$^{\mathrm{a}}$Figure \ref{attacked_model} Statistics}
        \end{tabular}
    \label{attacked_model_table}
    \end{table}
\end{center}

We investigate if we can find out if an attack occurred after we predicted the input [Figure \ref{attacked_model}, Table \ref{attacked_model_table}]. We take the class predictions from each of the models and attempt to use a random forest algorithm to predict if it came from an adversarial input or a clean input. This investigation will give us a clue about how we might safeguard production systems from an adversarial attack after they have already been implemented.  We show that this task is more than manageable with a mAP of around 66\% for both predicting a clean prediction versus an adversarial prediction. 

\subsection{\textbf{Model Attacked}}

\begin{figure}[t]
    \centering
    \includegraphics[width=0.5\textwidth]{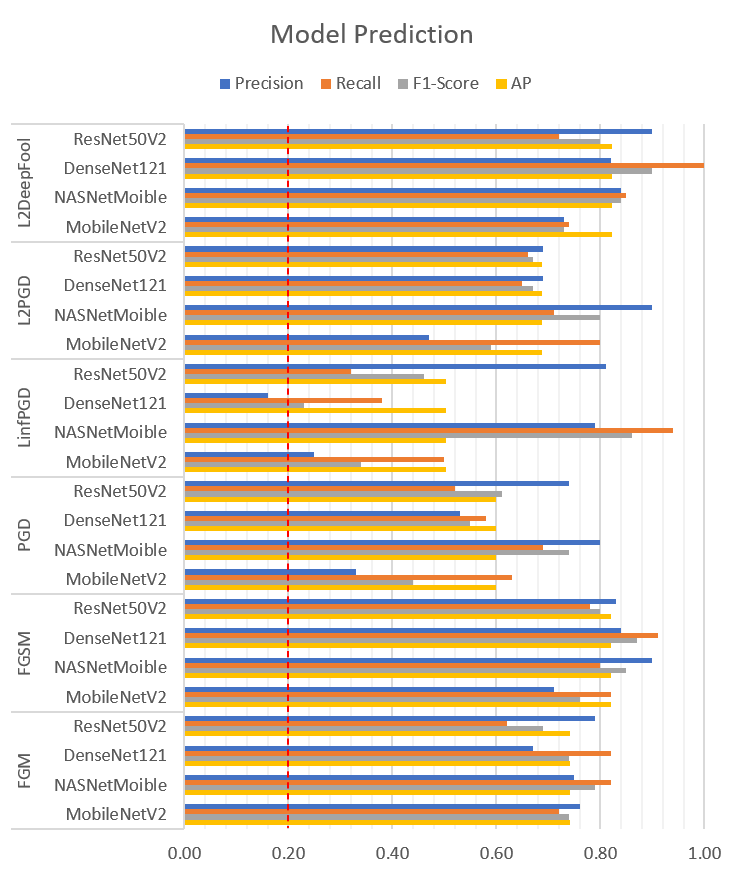}
    \caption{Model Attacked based on Model Prediction. Vertical dashed red line is random guessing.}
    \label{model_used}
\end{figure}

\begin{center}
    \begin{table}[t]
    \caption{Model Attacked from Class Prediction} 
    \centering
        \begin{tabular}{|c|c|c|c|}
            \hline
            \textbf{Model} & \textbf{mAP} & \textbf{Avg. Recall} & \textbf{Avg. F1} \\ \hline
            MobileNetV2	& 0.542	& 0.702	& 0.600\\ \hline
            NASNetMoible	& 0.830 & 0.802	& 0.813\\ \hline
            DenseNet121	& 0.618	& 0.723	& 0.660\\ \hline
            ResNet50V2	& 0.793	& 0.603	& 0.672\\ \hline
        \multicolumn{4}{l}{$^{\mathrm{a}}$Figure \ref{model_used} Statistics}
        \end{tabular}
        \label{model_used_table}
    \end{table}
\end{center}

When trying to predict, from the image class predictions [Figure \ref{model_used}, Table \ref{model_used_table}], which model was originally attack is possible using a random forest network. This is similar to the previous paper's result where the biases in the networks imprint unique non-linear patterns on the class predictions. We are able to recover that bias at almost 70\% AP.

\subsection{\textbf{Detect Adversarial Attack Type}}

\begin{figure}[t]
    \centering
    \includegraphics[width=0.5\textwidth]{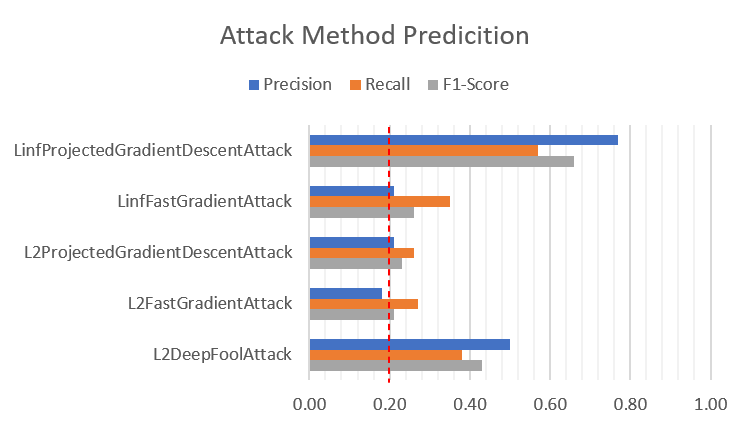}
    \caption{Attack Type from Model Prediction. Vertical dashed red line is random guessing.}
    \label{attack_type}
\end{figure}

\begin{center}
    \begin{table}[t]
    \caption{Attack Type from Model Prediction} 
    \centering
        \begin{tabular}{|c|c|c|c|}
            \hline
            \textbf{Attack} & \textbf{Precision} & \textbf{Recall} & \textbf{F1} \\ \hline
            L2DeepFool	& 0.50	& 0.38	& 0.43\\ \hline
            FGM	& 0.18	& 0.27	& 0.21\\ \hline
            L2PGD	& 0.21	& 0.26	& 0.23\\ \hline
            FGSM	& 0.21	& 0.35	& 0.26\\ \hline
            LinfPGD	& 0.77	& 0.57	& 0.66\\ \hline
        \multicolumn{4}{l}{$^{\mathrm{a}}$Figure \ref{attack_type} Statistics }
        \end{tabular}
        \label{attack_type_plot}
    \end{table}
\end{center}

After running each attack on each model, we attempt to decipher which attack was made by looking at a model output [Figure \ref{attack_type}, Table \ref{attack_type_plot}]. We expected that since each attack optimizes a different variable, that we could recover that bias at a high rate. However, we find that we can predict which attack type at 37\% AP (random guessing 20\%).

\subsection{ResNet50 Immunity} \label{resnet50_discussion}

The resnet50 model was resilient to all the white-box attacks from the foolbox package. This is a package error which involves how foolbox handles preprocessing images to send to models via the foolbox TensorFlowModel class.

\section{Discussion}
Adversarial Training focuses on using adversarial examples during the training step of model building. By detecting the incoming attacks, model builders can focus on tailoring their models to the most frequent attacks that are coming to their systems. Adversarial Training has gained increased popularity with new techniques reducing the amount of computational power necessary to make models more robust\cite{kurakin2016adversarial}. A recent paper, Using Single-Step Adversarial Training to Defend Iterative Adversarial Examples, proposed an iterative approach to introducing adversarial examples in increasing robustness of an underlying model\cite{liu2021using}. Using the detection methods proposed in this paper and an adversarial training pipeline, it is possible to protect a production system from the most frequent attacks.

\section{Future Work}
Detecting the incoming attacks threat vector allows a targeted system to protect and respond to these attacks. This paper introduces the attack detection method and a future work will focus on using these predictions in an attack early warning system. Each predicted attack will flow to a decision system that directs the inference engine to deflect, protect, or avoid the incoming request. Once an attacker is identified, we can revoke privileges or black list those attack venues.

\section{Conclusion}
We were successfully able to detect if a model was attack by varied adversarial attacks using only the model output. Not only were we able to detect if a model was attacked at 66\% AP, but also which model was attack at 70\% AP. Less impressive is classifying which attack type was used at 37\% AP. In an adversarial defense system, we envision the routing of inference through an adversarial detection system to help avoid negative downstream consequences. In addition, we show a detection model could be trained with only a handful of adversarial examples and can run in less than a millisecond (0.12ms).

\section*{Acknowledgment}
The authors would like to thank Auburn University and the PeopleTec Technical Fellows program for encouragement and project assistance. 

\bibliographystyle{./IEEEtran.bst}
\bibliography{./refs.bib}

\begin{thebibliography}{10}
\providecommand{\url}[1]{#1}
\csname url@samestyle\endcsname
\providecommand{\newblock}{\relax}
\providecommand{\bibinfo}[2]{#2}
\providecommand{\BIBentrySTDinterwordspacing}{\spaceskip=0pt\relax}
\providecommand{\BIBentryALTinterwordstretchfactor}{4}
\providecommand{\BIBentryALTinterwordspacing}{\spaceskip=\fontdimen2\font plus
\BIBentryALTinterwordstretchfactor\fontdimen3\font minus
  \fontdimen4\font\relax}
\providecommand{\BIBforeignlanguage}[2]{{%
\expandafter\ifx\csname l@#1\endcsname\relax
\typeout{** WARNING: IEEEtran.bst: No hyphenation pattern has been}%
\typeout{** loaded for the language `#1'. Using the pattern for}%
\typeout{** the default language instead.}%
\else
\language=\csname l@#1\endcsname
\fi
#2}}
\providecommand{\BIBdecl}{\relax}
\BIBdecl

\bibitem{kalin2020black}
J.~Kalin, M.~Ciolino, D.~Noever, and G.~Dozier, ``Black box to white box:
  Discover model characteristics based on strategic probing,'' \emph{arXiv
  preprint arXiv:2009.03136}, 2020.

\bibitem{huang2011adversarial}
L.~Huang, A.~D. Joseph, B.~Nelson, B.~I. Rubinstein, and J.~D. Tygar,
  ``Adversarial machine learning,'' in \emph{Proceedings of the 4th ACM
  workshop on Security and artificial intelligence}, 2011, pp. 43--58.

\bibitem{machado2020adversarial}
G.~R. Machado, E.~Silva, and R.~R. Goldschmidt, ``Adversarial machine learning
  in image classification: A survey towards the defender's perspective,''
  \emph{arXiv preprint arXiv:2009.03728}, 2020.

\bibitem{chakraborty2018adversarial}
A.~Chakraborty, M.~Alam, V.~Dey, A.~Chattopadhyay, and D.~Mukhopadhyay,
  ``Adversarial attacks and defences: A survey,'' \emph{arXiv preprint
  arXiv:1810.00069}, 2018.

\bibitem{xu2020adversarial}
H.~Xu, Y.~Ma, H.-C. Liu, D.~Deb, H.~Liu, J.-L. Tang, and A.~K. Jain,
  ``Adversarial attacks and defenses in images, graphs and text: A review,''
  \emph{International Journal of Automation and Computing}, vol.~17, no.~2, pp.
  151--178, 2020.

\bibitem{wang2019security}
X.~Wang, J.~Li, X.~Kuang, Y.-a. Tan, and J.~Li, ``The security of machine
  learning in an adversarial setting: A survey,'' \emph{Journal of Parallel and
  Distributed Computing}, vol. 130, pp. 12--23, 2019.

\bibitem{shan2019using}
S.~Shan, E.~Wenger, B.~Wang, B.~Li, H.~Zheng, and B.~Y. Zhao, ``Using honeypots
  to catch adversarial attacks on neural networks,'' \emph{arXiv preprint
  arXiv:1904.08554}, 2019.

\bibitem{metzen2017detecting}
J.~H. Metzen, T.~Genewein, V.~Fischer, and B.~Bischoff, ``On detecting
  adversarial perturbations,'' \emph{arXiv preprint arXiv:1702.04267}, 2017.

\bibitem{reproducible_colab}
\BIBentryALTinterwordspacing
M.~Ciolino, ``Google collaboratory reproducible image experiment,'' in
  \emph{2021}, 2021. [Online]. Available:
  \url{https://colab.research.google.com/drive/1MQsjpra1EWo41gMPabh-arxlrfPhvrVw?usp=sharing}
\BIBentrySTDinterwordspacing

\bibitem{recht2019imagenet}
B.~Recht, R.~Roelofs, L.~Schmidt, and V.~Shankar, ``Do imagenet classifiers
  generalize to imagenet?'' in \emph{International Conference on Machine
  Learning}.\hskip 1em plus 0.5em minus 0.4em\relax PMLR, 2019, pp. 5389--5400.

\bibitem{scikit-learn}
F.~Pedregosa, G.~Varoquaux, A.~Gramfort, V.~Michel, B.~Thirion, O.~Grisel,
  M.~Blondel, P.~Prettenhofer, R.~Weiss, V.~Dubourg, J.~Vanderplas, A.~Passos,
  D.~Cournapeau, M.~Brucher, M.~Perrot, and E.~Duchesnay, ``Scikit-learn:
  Machine learning in {P}ython,'' \emph{Journal of Machine Learning Research},
  vol.~12, pp. 2825--2830, 2011.

\bibitem{rauber2017foolboxnative}
\BIBentryALTinterwordspacing
J.~Rauber, R.~Zimmermann, M.~Bethge, and W.~Brendel, ``Foolbox native: Fast
  adversarial attacks to benchmark the robustness of machine learning models in
  pytorch, tensorflow, and jax,'' \emph{Journal of Open Source Software},
  vol.~5, no.~53, p. 2607, 2020. [Online]. Available:
  \url{https://doi.org/10.21105/joss.02607}
\BIBentrySTDinterwordspacing

\bibitem{rauber2017foolbox}
J.~Rauber, W.~Brendel, and M.~Bethge, ``Foolbox: A python toolbox to benchmark
  the robustness of machine learning models,'' \emph{arXiv preprint
  arXiv:1707.04131}, 2017.

\bibitem{art2018}
\BIBentryALTinterwordspacing
M.-I. Nicolae, M.~Sinn, M.~N. Tran, B.~Buesser, A.~Rawat, M.~Wistuba,
  V.~Zantedeschi, N.~Baracaldo, B.~Chen, H.~Ludwig, I.~Molloy, and B.~Edwards,
  ``Adversarial robustness toolbox v1.2.0,'' \emph{CoRR}, vol. 1807.01069,
  2018. [Online]. Available: \url{https://arxiv.org/pdf/1807.01069}
\BIBentrySTDinterwordspacing

\bibitem{madry2017towards}
A.~Madry, A.~Makelov, L.~Schmidt, D.~Tsipras, and A.~Vladu, ``Towards deep
  learning models resistant to adversarial attacks,'' \emph{arXiv preprint
  arXiv:1706.06083}, 2017.

\bibitem{kurakin2016adversarial}
A.~Kurakin, I.~Goodfellow, and S.~Bengio, ``Adversarial machine learning at
  scale,'' \emph{arXiv preprint arXiv:1611.01236}, 2016.

\bibitem{liu2021using}
G.~Liu, I.~Khalil, and A.~Khreishah, ``Using single-step adversarial training
  to defend iterative adversarial examples,'' in \emph{Proceedings of the
  Eleventh ACM Conference on Data and Application Security and Privacy}, 2021,
  pp. 17--27.

\end{thebibliography}

\end{document}